\documentclass{article}

\PassOptionsToPackage{numbers, compress, square}{natbib}

\usepackage[preprint]{neurips_2024}

\usepackage[utf8]{inputenc} 
\usepackage[T1]{fontenc}    
\usepackage{hyperref}       
\usepackage{url}            
\usepackage{booktabs}       
\usepackage{amsfonts}       
\usepackage{nicefrac}       
\usepackage{microtype}      
\usepackage{xcolor}         
\usepackage{graphicx}       
\usepackage{amssymb}        
\usepackage{algorithm}
\usepackage{algorithmic}
\usepackage{amsmath} 
\usepackage{bm}

\usepackage{enumerate,enumitem}

\title{Controllable Continual Test-Time Adaptation}

\author{Ziqi Shi$^{12}$\thanks{Equal contribution} \quad Fan Lyu$^{1*}$ \quad Ye Liu$^{3}$ \quad Fanhua Shang$^{2}$ \\
\quad \textbf{Fuyuan Hu}$^{4}$  \quad \textbf{Wei Feng}$^{2}$ \quad \textbf{Zhang Zhang}$^{1}$\thanks{Corresponding author.}  \quad \textbf{Liang Wang}$^{1}$ \\
$^1$New Laboratory of Pattern Recognition, Institute of Automation, Chinese Academy of Sciences \\ 
$^2$College of Intelligence and Computing, Tianjin University \\
$^3$School of Computer Science \& Engineering, Linyi University \\
$^4$School of Electronic \& Information Engineering, Suzhou University of Science and Technology \\
\texttt{\{szqi, fhshang, wfeng\}@tju.edu.cn} \
\texttt{\{zzhang, wangliang\}@nlpr.ia.ac.cn} \\
\texttt{fan.lyu@cripac.ia.ac.cn} \
\texttt{liuye21@lyu.edu.cn} \
\texttt{fuyuanhu@mail.usts.edu.cn} \\
}

\begin{document}

\maketitle

\begin{abstract}

Continual Test-Time Adaptation (CTTA) is an emerging and challenging task where a model trained in a source domain must adapt to continuously changing conditions during testing, without access to the original source data. CTTA is prone to error accumulation due to uncontrollable domain shifts, leading to blurred decision boundaries between categories. Existing CTTA methods primarily focus on suppressing domain shifts, which proves inadequate during the unsupervised test phase.
In contrast, we introduce a novel approach that guides rather than suppresses these shifts.
Specifically, we propose \textbf{C}ontrollable \textbf{Co}ntinual \textbf{T}est-\textbf{T}ime \textbf{A}daptation (C-CoTTA), which explicitly prevents any single category from encroaching on others, thereby mitigating the mutual influence between categories caused by uncontrollable shifts. 
Moreover, our method reduces the sensitivity of model to domain transformations, thereby minimizing the magnitude of category shifts. 
Extensive quantitative experiments demonstrate the effectiveness of our method, while qualitative analyses, such as t-SNE plots, confirm the theoretical validity of our approach. Our code is available at \href{https://github.com/RenshengJi/C-CoTTA}{https://github.com/RenshengJi/C-CoTTA.}

\end{abstract}

\section{Introduction}

Continual Test-Time Adaptation (CTTA) \cite{wang2022continual} is becoming an emerging field, which explores the adaptability of any machine learning model during test time in dynamic environments. 
The primary objective of CTTA is to enable a pretrained model to adapt to continuously changing scenarios, where the distribution of data shifts over time.
CTTA is practical in many long-term intelligent applications, such as autopilot \cite{hu2022sim, o2021accelerated, chen2020domain}, monitoring \cite{geiger2013vision}, medical image analysis \cite{chen2023each, gonzalez2020wrong}, where models need to remain robust and accurate against possible changes over extended periods.

\begin{figure}[t]
    \centering
    \includegraphics[width=.9\linewidth]{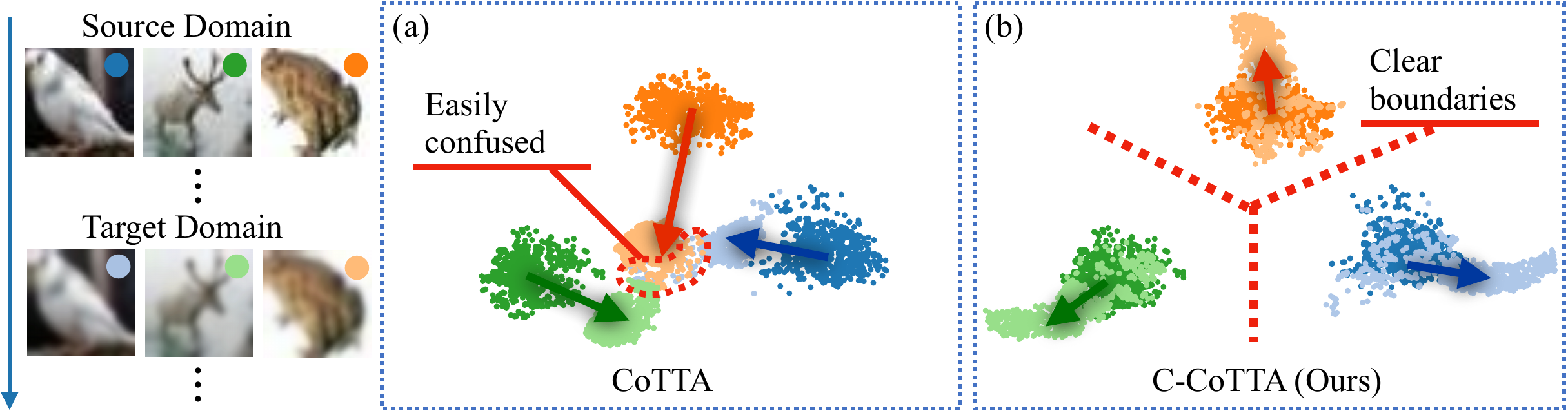}
    \vspace{-5px}
    \caption{t-SNE visualization of controllable domain shift in CTTA. (a) For CoTTA~\cite{wang2022continual}, due to the lack of control over domain shift, categories being biased towards others, resulting in fuzzy classification boundaries. (b) In contrast, our method achieves controllable domain shift, so even if categories are shift, it will not lead to confusion among categories.}
    \label{fig:case}
    \vspace{-10px}
\end{figure}

The main challenge of CTTA is the accumulation of errors caused by the lack of real labels and the continuous domain shifts. 
Existing methods mainly avoid error accumulation by using strategies such as Mean Teacher \cite{tarvainen2017mean}, augmentation-averaged predictions \cite{wang2022continual, lyu2024variational, liu2023vida}, and selecting reliable samples \cite{yang2023exploring, niloy2024effective, niu2022eata, wang2024continual} to minimize the impact of domain shift. 
However, these methods focus on suppress the shifts, few methods explicitly attempt to guide or control the shifts.
This is becuase the lack of labels in the target domain and the inability to obtain source domain data in CTTA. 
We evaluate this on CIFAR10C, as shown in Fig.~\ref{fig:case}(a), we find that in the domain adaptation process, multiple categories to lean towards a confusion region.
In fact, CTTA meets continuous unknown domain shifts, making this shift almost uncontrollable. 
This phenomenon will worsen with the changing domain over time, resulting in error accumulation.

Therefore, \textit{we hypothesize that instead of attempting to suppress the shift, guiding or controlling the shift to remain class-separable may also achieve effective adaptation}.
On top of this, the main focus of this paper is to study how to achieve controllable domain shift in CTTA. 
Generally, controlling domain shifts means to control the direction of the shift in feature space. 
First of all, we first need to accurately represent the direction of domain shift. 
Inspired by the interpretable machine learning~\cite{kim2018interpretability}, we represent shift directions using the tool of Concept Ativation Vectors (CAV)~\cite{pahde2022patclarc}, which represents the transformation path from one concepts to another.
In CTTA, the CAV can be represented by the vector from one prototype to another. 
First, for a specific category of domain shift direction, we obtain it by subtracting the category prototype of the target domain in the feature space from the category prototype of the source domain. Then, for the overall domain shift direction, we obtain it by subtracting the overall prototype of the target domain in the feature space from the overall prototype of the source domain.

To further control domain shift, we construct the Control Domain Shift (CDS) loss and the Control Class Shift (CCS) loss.
CDS refers to controlling the shift of the overall domain by constraining the model's sensitivity in that direction, thus reducing the impact of domain shift on model performance. 
CCS controls the shift of specific categories by constraining the shift direction of each category to avoid biasing other categories. 
As shown in \text{Fig.}~\ref{fig:case}(b), our method achieve controllable domain shift, and the direction of the shift will not blur the classification boundaries.
Extensive experiments are conducted on three large-scale benchmark datasets to validate the effectiveness of the proposed C-CoTTA framework in various challenging and realistic scenarios.

Our contributions are three-fold:
\begin{enumerate}[label=(\arabic*),left=0pt,itemsep=0pt]
    \vspace{-5px}
    \item We evaluate and find that only suppress domain shifts is insufficient, which may lead to blurred classification boundaries. In contrast, we propose to guide and control shifts to keep the class-separability.
    \item We propose a simple and effective direction representation based on Concept Ativation Vectors (CAV) in interpretable machine learning, which utilizes the difference between two prototypes in the feature space.
    \item We propose to explicitly control the direction of specific category bias by preventing any category from leaning towards other categories, in order to prevent the blurring of classification boundaries; at the same time, by reducing the sensitivity of the model to domain shift, we control the overall domain shift to alleviate the impact of domain shift on domain adaptation.
\end{enumerate}

\section{Related work}
\subsection{Continual Test-Time Adaptation}

Continual Test-Time Adaptation (CTTA) \cite{wang2022continual, lyu2024elastic, lyu2021multi} is an emerging paradigm within the machine learning community designed to address the dynamic nature of real-world data distributions. Unlike traditional Test-Time Adaptation (TTA) \cite{jain2011online, sun2020test, wang2020tent}, which typically assumes a fixed target domain in a source-free and online manner, CTTA operates under the assumption that the target domain may change over time. The main challenge in CTTA is the potential for catastrophic forgetting and error accumulation. During the test time, as the model adapts to new distributions, it risks losing previously learned knowledge, which can result in a degradation in performance known as catastrophic forgetting \cite{van2019three}. Moreover, the utilization of pseudo-labels derived from the model's own predictions can introduce errors, which may accumulate over time \cite{li2017learning, wang2022continual}, especially when there are frequent domain shifts.

To address the challenges of error accumulation in CTTA, researchers have developed various strategies. 
A number of works \cite{wang2022continual, lyu2024variational, liu2023vida, du2023multi} employ augmentation-averaged predictions for the teacher model to boost the teacher's confidence, while others\cite{chakrabarty2023sata, dobler2023robust} add perturbations to the student to enhance the model's robustness. 
Various methods\cite{yang2023exploring, niloy2024effective, niu2022eata, wang2024continual, sun2022exploring} focuses on selecting reliable samples to eliminate the impact of misclassified samples on domain adaptation. 
As to the challenge of catastrophic forgetting, Wang et al. \cite{wang2022continual} and Brahma et al.\cite{brahma2023probabilistic}
believe that the source model is more reliable, thus they are designed to restore the source parameters.
While these studies address the CTTA issue at the model level, other research efforts \citep{gan2023decorate, yang2023exploring, ni2023distribution} leverage visual domain prompts or a limited subset of parameters to extract ongoing target domain knowledge.
However, these approaches primarily focus on suppressing domain shift and there are few methods that explicitly attempt to guide or control domain shift.

\subsection{Concept Activation Vectors}
Concept Activation Vectors (CAVs) is an interpretability tool for explaining decision-making processes in deep learning models. Originally, the authors of \cite{kim2018interpretability} define CAV as the normal to a hyperplane that separates examples without a concept from examples with a concept in the model's latent activations. This hyperplane is commonly computed by solving a classification problem, for example, using Support Vector Machines (SVMs)\cite{anders2022finding}, ridge\cite{cortes1995support}, lasso\cite{pfau2021robust} or logistic regression \cite{yuksekgonul2022post}.
Given its ability to effectively orient concepts, CAVs have been employed for a plethora of tasks in recent years, such as concept sensitivity testing
~\cite{kim2018interpretability}, model correction for shortcut removal ~\cite{anders2022finding,pahde2023reveal,dreyer2023hope}, 
knowledge discovery by investigation of internal model states~\cite{mcgrath2022acquisition}, 
and training of post-hoc concept bottleneck models~\cite{yuksekgonul2022post}.
However, common regression-based methods tend to deviate from the true conceptual direction due to factors such as noise in the data \cite{haufe2014interpretation}. To that end, signal-pattern-based CAVs (referred to “signalCAVs”) have been proposed \cite{pahde2022patclarc}, which are more robust against noise \cite{weber2023beyond, dreyer2024hope, samek2023explainable, biecek2024explain}.
However, during test-time, we may not have access to all samples of a prototype, and due to the lack of true labels, misclassified samples may contaminate the prototype. Therefore, the construction of a prototype is different from interpretable machine learning.

\section{Methodology}
\newcommand{\cov}[2]{\textrm{cov}[#1,#2]}

\subsection{Problem Definition}

Given a classification model pre-trained on a source domain, CTTA methods adapt the source model to the unlabeled target data, where the domain continuously changes.
The unsupervised dataset of target domains are denoted as $\mathcal{D}^k = \{x^k_m\}^{N^k}_{m=1}$, where $k$ is the target domain index.
As shown in \text{Figure}~\ref{fig:case}, in the process of CTTA, if the domain shifts are not controlled, some categories may generate bias towards other categories, resulting in blurred classification boundaries. 
In this paper, we propose to explicit control over the shift direction in CTTA.
specific categories and the overall domain.
In the following, we first study how to represent domain shifts in Sec.~\ref{sec:rep}.
Then, we propose to control the shift within the process of CTTA in Sec.~\ref{sec:ctrl}.

\subsection{Representing Shift via CAV}

\label{sec:rep}

Domain shift refers to the distribution shift of each class that occurs in the feature space, generally caused by differences between the target domain distribution in the testing phase and the source domain distribution in the training phase. 
Therefore, how to represent the domain shift direction in the feature space in CTTA is a challenge.

In the field of interpretable machine learning~\cite{kim2018interpretability, mcgrath2022acquisition}, CAV~\cite{kim2018interpretability} refer to the normal to a hyperplane that separates examples without a concept from examples with a concept in the model's latent activations feature space. 
CAV is widely used in areas such as model correction for shortcut removal~\cite{anders2022finding,pahde2023reveal,dreyer2023hope}. 
The concept in CAV generally refers to high-level semantic information, such as whether there are a large number of striped structures in an image. 
For example, in the concept of stripe, the label for features extracted from images with stripes is 1, and without stripes is 0. 

CAV can be calculated in different methods, we use the signal-pattern-based CAVs (SCAV) \cite{pahde2022patclarc} which provide a simple but effective way to represent the CAV $\mathbf{v}$ as follows:
\begin{equation}\label{eq:30}
\mathbf{v} =\frac{\cov{f(x)}{t}}{\cov{t}{t}} = \frac{\sum (f(x_{i})-\overline{f}(x))(t_{i}-\overline{t})}{\sum(t_i-\overline{t})^2} , \quad t_{i} =
\begin{cases}
1 & \text{if } x_{i} \in \mathcal{X}_{c}, \\
0 & \text{if } x_{i} \in \mathcal{X}_{n}.
\end{cases}
,
\end{equation}
where $f(\cdot)$ represents the feature extractor and $t$ represents the concept label of the features, $\mathcal{X}_{c}$ and $\mathcal{X}_{n}$ denote sets of samples that either possess or lack the specified concept, respectively. 
Mean feature $\overline{f}(x)=\frac{1}{N}\sum f(x_{i})$ and mean label $\overline{t}=\frac{1}{N}\sum t_{i}$.

\begin{figure}[t]
    \centering
    \includegraphics[width=.9\linewidth]{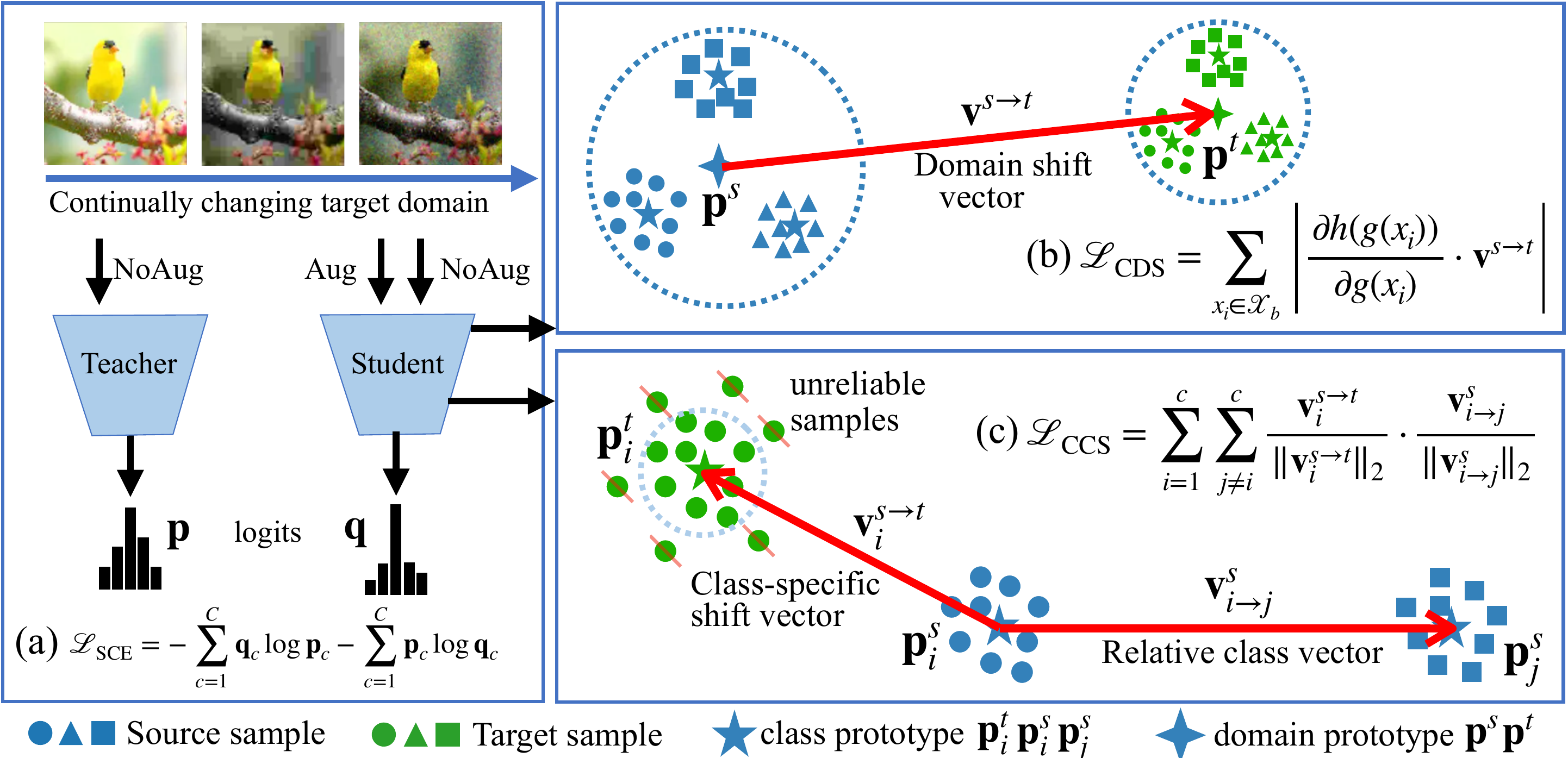}
    \caption{The pipeline of C-CoTTA. (a) Based on the mean teacher framework, perturb the student to enhance model robustness, while optimizing using symmetric cross-entropy. (b) Control the overall domain shift by constraining the model's sensitivity to domain shift directions. (c) Control the shift of a specific category by directly controlling the shift direction of any category to prevent bias towards other categories.}
    \label{fig:pipeline}
    \vspace{-10px}
\end{figure}

Nevertheless, in the traditional CAV application scenario, the concept label $t$ is manually annotated offline. In contrast, in unsupervised CTTA scenario, to represent the domain shift direction in an online manner is intractable.
Therefore, we use the pseudo-label $y$ obtained by the existing model to represent concepts automatically. 
The shift representation for the direction in CTTA can be computed as follows:
\begin{equation}\label{eq:1}
\mathbf{v} =\frac{\cov{f(x)}{y}}{\cov{y}{y}} = \frac{\sum (f(x_{i})-\overline{f}(x))(y_{i}-\overline{y})}{\sum(y_i-\overline{y})^2} , \quad y_{i} =
\begin{cases}
1 & \text{if } x_{i} \in \mathcal{X}_{t}, \\
0 & \text{if } x_{i} \in \mathcal{X}_{s}.
\end{cases}
,
\end{equation}
At the same time, through deduction (refer to Appendix~\ref{sec:appendix1} for details), it is determined that the direction can be further represented as the difference between two prototypes in the feature space.
\begin{equation}\label{eq:2}
\mathbf{v}=\frac{1}{|\mathcal{X}_{t}|}{\sum\limits_{x_{i} \in \mathcal{X}_{t}} f(x_i)} - \frac{1}{|\mathcal{X}_{s}|}{\sum\limits_{x_{i} \in \mathcal{X}_{s}} f(x_i)} = \mathbf{p}_{t}- \mathbf{p}_{s}.
\end{equation}
Then, we can use this to calculate various directional vectors, in order to further control domain shift. 

\textbf{Domain-level source-to-target shift.}
We represent the shift of the overall domain as below:
\begin{equation}\label{eq:22}
\mathbf{v}^{s \to t} = \mathbf{p}^t - \mathbf{p}^s,
\end{equation} 
This calculates by subtracting the target domain prototype $\mathbf{p}^t$ composed of all samples in each batch at test-time from the source domain prototype $\mathbf{p}^s$ composed of prototypes of all categories in the source domain as shown in Figure~\ref{fig:pipeline}(b).

\textbf{Class-level source-to-target shift.}
We also represent the shift of a specific category as below:
\begin{equation}\label{eq:20}
\mathbf{v}_{i}^{s \to t} = \mathbf{p}^t_i - \mathbf{p}^s_i,
\end{equation} where we construct it by subtracting the prototype $\mathbf{p}^t_i$ composed of trustworthy samples belonging to that category in each batch at test-time from the source domain prototype $\mathbf{p}^s_i$ of that category as shown in Figure~\ref{fig:pipeline}(c). Moreover, during test time, some samples may be misclassified due to the lack of real labels, leading to distortion in the extracted prototypes. Therefore, we estimate the entropy $H(x^t)$ for each sample $x^t$ from the target domain using the model. We then set aside samples with an entropy exceeding a predefined threshold $E_0$ following \cite{niu2022eata}.

\textbf{Class-to-class source shift.}
We construct class relative shift from category $i$ to $j$ as below:
\begin{equation}\label{eq:21}
\mathbf{v}_{i \to j}^{s} = \mathbf{p}^s_j - \mathbf{p}^s_i.
\end{equation}  
This calculates by using the source domain category prototypes $\mathbf{p}^s_i$ and $\mathbf{p}^s_j$ as shown in Figure~\ref{fig:pipeline}(c).

\subsection{Control of shift}

\label{sec:ctrl}
We control the shift at domain and class levels. 
On one hand, the shift of different categories are influenced by the overall domain shift, showing a certain degree of convergence, meaning that after domain shift, the distances between different categories in the feature space become closer. This finding is similar to the conclusions drawn in studies such as ~\cite{xu2019larger, kondo2022source, rahman2021deep}, where researchers found that the feature norms in the target domain are relatively small, so we control the overall domain shift. 
On the other hand, the shift of different categories is related to their characteristics.

\paragraph{Domain-level shift controlling.}
First, we propose to control the shift of the overall domain at the domain level. Specifically, we model the shift sensitivity by considering the gradient of the model output $h(g(x_{i}))$ with respect to the feature map $g(x_{i})$, and then combine it with the overall domain shift direction ${\mathbf{v}^{s\to t}}$. We can then reduce the sensitivity of the model to domain shift by constraining the direction gradient as follows:
\begin{align}\label{eq:7}
\mathcal{L}_{\mathrm{CDS}}=\sum\limits_{x_{i} \in {\mathcal{X}_b}}\left|\frac{\partial h(g(x_{i}))}{\partial g(x_{i})}\cdot {\mathbf{v}^{s\to t}} \right|,
\end{align}
where $h(\cdot)$ represents the remaining part of the model. Intuitively, the Control Domain Shift (CDS) loss $\mathcal{L}_{\mathrm{CDS}}$ enforces the model output to not change when slightly adding or removing activations along the bias direction as follows:
\begin{align}\label{eq:8}
\lim _{\epsilon \rightarrow 0} \frac{h(g(\mathbf{x})+\epsilon {\mathbf{v}^{s\to t}})-h(g(\mathbf{x}))}{\epsilon} = 0.
\end{align}
Thus, by minimizing $\mathcal{L}_{\mathrm{CDS}}$, the model becomes insensitive towards the domain shift direction, thereby reducing the impact of domain shift on the domain adaptation process.

\begin{algorithm}[t]
    \caption{Controllable Continual Test-Time Adaptation}
    \label{alg:ccotta}
    \begin{algorithmic}[1]
    \REQUIRE{Target domains data $\mathcal{D}^k = \{x^k_m\}^{N^k}_{m=1}$, Source model, Source domain class prototype $\mathbf{p}_i^s$}
    \STATE Generate the category relative direction vectors $\mathbf{v}_{i \to j}^{s} = \mathbf{p}^s_j - \mathbf{p}^s_i$ before domain adaptation
    \FOR{a domain $k$ in $K$}
        \FOR{a batch $\{x^k_b\}_{b=1}^{B}$ in $\mathcal{D}^k$}
        \STATE Forward the batch, make predictions and get features
        \STATE Identify reliable samples with low entropy using a predefined threshold $E_0$
        \STATE Compute the direction of the domain shift $\mathbf{v}^{s \to t} = \mathbf{p}^t - \mathbf{p}^s$ and constrained via \text{Eq.}~\ref{eq:7}
        \STATE Compute the direction of the class shift $\mathbf{v}_{i}^{s \to t} = \mathbf{p}^t_i - \mathbf{p}^s_i$ and constrained via \text{Eq.}~\ref{eq:5}
        \STATE Compute the symmetric cross-entropy loss via \text{Eq.}~\ref{eq:10}
        \STATE Optimize model by minimizing $\mathcal{L}$ via \text{Eq.}~\ref{eq:9} and update student and teacher models
        \ENDFOR
    \ENDFOR
    \end{algorithmic}
\end{algorithm}

\paragraph{Class-level shift controlling.}
Then, in order to prevent uncontrollable shifts of each category, 
we propose to control class-level shift to avoid any category leaning towards other categories. 
This requires that the direction of shift $\mathbf{v}_{i}^{s \to t}$ for any category is preferably the opposite direction of the direction $\mathbf{v}_{i \to j}^{s}$ of other categories relative to that category as shown in \text{Fig.}~\ref{fig:pipeline}(c).
This means that the dot product of $\mathbf{v}_{i}^{s \to t}$ and $\mathbf{v}_{i \to j}^{s}$ should be as small as possible. The Control Class Shift (CCS) loss is calculated as follows:
\begin{align}\label{eq:5}
\mathcal{L}_{\mathrm{CCS}}=\sum_{i=1}^{c}\sum_{j \neq i}^{c}\frac{\mathbf{v}_{i}^{s \to t}}{\|{\mathbf{v}_{i}^{s \to t}}\|_{2}}\cdot\frac{{\mathbf{v}_{i\to j}^s}}{\|{\mathbf{v}_{i\to j}^s}\|_{2}},
\end{align}
where we normalize $\mathbf{v}_{i}^{s \to t}$ and $\mathbf{v}_{i \to j}^{s}$.
The loss $\mathcal{L}_{\mathrm{CCS}}$ is to prevent any category from shifting towards other categories, achieving controllable domain shift, effectively preventing the decrease in classification performance caused by blurred category boundaries in continuous domain adaptation.

\subsection{Overall Objective}

In addition, our work also utilizes $\mathcal{L}_{\mathrm{SCE}}$ derived from \cite{dobler2023robust}, which following the mean teacher framework, which involves simply averaging the weights of a student model over time, the resulting teacher model provides a more accurate prediction function than the final function of the student, meanwhile perturbs the student to enhance the robustness of the model \cite{xie2020self, sohn2020fixmatch} and employs the symmetric cross-entropy (SCE) \cite{wang2019symmetric} which has superior gradient properties compared to the commonly used cross-entropy loss. For enhancing the output of students $q$ and teachers $p$, the symmetric cross-entropy is defined as follows: 
\begin{align}\label{eq:10}
    \mathcal{L}_{\mathrm{SCE}} = - \sum\nolimits_{c=1}^C \mathbf{q}_c\, \mathrm{log}\, \mathbf{p}_c - \sum\nolimits_{c=1}^C \mathbf{p}_c\, \mathrm{log}\, \mathbf{q}_c.
\end{align}

The overall objective of our proposed continual test-time adaptation method is as follows:
\begin{align}\label{eq:9}
\mathcal{L}= \mathcal{L}_{\mathrm{SCE}} + \lambda_1\mathcal{L}_{\mathrm{CDS}} + \lambda_2\mathcal{L}_{\mathrm{CCS}},
\end{align}
where $\lambda_1$, and $\lambda_2$ are the hyperparameters. $\mathcal{L}_{\mathrm{CDS}}$ refers to controlling the shift of the overall domain, which constrains the model's sensitivity in that direction. $\mathcal{L}_{\mathrm{CCS}}$ controls the shift of specific categories, which constrains the shift direction of each category to avoid biasing other categories. The overall frame diagram is shown in \text{Figure}~\ref{fig:pipeline}.

We illustrate the whole algorithm in Algorithm~\ref{alg:ccotta}. First, before domain adaptation begins, we use the source domain category prototypes to calculate the inter-class relative direction vector $\mathbf{v}_{i \to j}^{s}$. During domain adaptation, on one hand, we calculate the shift direction $\mathbf{v}_{i}^{s \to t}$ for specific categories and constrain it through the loss $\mathcal{L}_{\mathrm{CCS}}$; on the other hand, we calculate the shift direction $\mathbf{v}^{s \to t}$ for the entire domain and constrain it through the loss $\mathcal{L}_{\mathrm{CDS}}$. Additionally, we compute the symmetric cross-entropy loss for the prediction logits of the student and teacher, optimize it  via \text{Eq.}~\ref{eq:9}, and update the student and teacher models.

\section{Experiments}

\subsection{Experimental Setting}
\textbf{Datasets.} 
We evaluate our proposed method on three continual test-time adaptation benchmark: CIFAR10-C, CIFAR100-C, and ImageNet-C. Each dataset contains 15 types of corruptions with 5 levels of severity, ranging from 1 to 5.  For simplicity in tables, we use Gauss., Impul., Defoc., Brit., Contr., Elas. and Pix. to represent Gaussian, Impulse, Defocus, Brightness, Contrast, Elastic,and Pixelate, respectively.

\textbf{Pretrained Model.} 
Following previous studies~\cite{wang2020tent,wang2022continual}, we adopt the pretrained WideResNet-28~\cite{zagoruyko2016wide}, ResNeXt-29~\cite{xie2017aggregated} and ResNet-50~\cite{he2016deep} for CIFAR10-C, CIFAR100-C and Imagenet-C, respectively. Similar to CoTTA, we update all the trainable parameters in all experiments. 

\textbf{Methods to be Compared.}
We compare our C-CoTTA with the original model (Source) and multiple state-of-the-art (SOTA) methods such as BN~\cite{li2017learning,schneider2020improving}, TENT~\cite{wang2020tent}, CoTTA~\cite{wang2022continual}, RoTTA~\cite{yuan2023robust}, SATA~\cite{chakrabarty2023sata}, SWA~\cite{yang2023exploring}, PETAL~\cite{brahma2023probabilistic}, RMT~\cite{dobler2023robust}, DSS~\cite{wang2024continual}. All compared methods utilize the same backbone and pretrained model.
All experiments are conducted on a single RTX 4090.

\subsection{Results for Continual Test-Time Adaptation}

\textbf{Experiments on CIFAR10-C.}
We first evaluate the effectiveness of the proposed model on the CIFAR10-C dataset. We compare our method to the source-only baseline and nine SOTA methods. 
As shown in Table~\ref{cifar10c}, directly using pre-trained model without adaptation yields a high average error rate of 43.5\%. BN method improve the performance by 23.1\% compared to the source-only baseline.
Among all comparison methods, SWA achieve the lowest error rate of 11.2\%, 11.8\% and 11.5\% on Defoc., Motion and Fog, respectively. Both PETAL and SATA achieve the lowest error rate of 7.4\% on Brit.. In other conditions, our proposed method outperforms or is comparable to all the above methods. In conclusion, our method achieve the lowest average error rate, which is reduced to 14.7\%.

\begin{table}[t]
    \setlength{\belowcaptionskip}{-0.2cm}
    \caption{Classification error rate (\%) for standard CIFAR10-C continual test-time adaptation task.}
    \label{cifar10c}
    \centering
    \resizebox{\textwidth}{!}{%
\setlength\tabcolsep{3pt}
\begin{tabular}{@{}lcccccccccccccccc@{}}
\toprule
Method & Gauss. & Shot & Impul. & Defoc. & Glass & Motion & Zoom & Snow & Frost & Fog  & Brit. & Contr. & Elas. & Pix. & Jpeg & Mean \\ 
\midrule
Source & 72.3 & 65.7 & 72.9 & 46.9 & 54.3 & 34.8 & 42.0 & 25.1 & 41.3 & 26.0 & \textcolor{white}{0}9.3  & 46.7 & 26.6 & 58.5 & 30.3 & 43.5 \\
BN~\cite{li2017learning,schneider2020improving}     & 28.1 & 26.1 & 36.3 & 12.8 & 35.3 & 14.2 & 12.1 & 17.3 & 17.4 & 15.3 & \textcolor{white}{0}8.4  & 12.6 & 23.8 & 19.7 & 27.3 & 20.4 \\
TENT~\cite{wang2020tent}   & 24.8 & 20.6 & 28.5 & 15.1 & 31.7 & 17.0 & 15.6 & 18.3 & 18.3 & 18.1 & 11.0 & 16.8 & 23.9 & 18.6 & 23.9 & 20.1 \\
CoTTA~\cite{wang2022continual}  & 24.5 & 21.5 & 25.9 & 12.0 & 27.7 & 12.2 & 10.7 & 15.0 & 14.1 & 12.7 & \textcolor{white}{0}7.6  & 11.0 & 18.5 & 13.6 & 17.7 & 16.3 \\
RoTTA ~\cite{yuan2023robust} & 30.3 & 25.4 & 34.6 & 18.3 & 34.0 & 14.7 & 11.0 & 16.4 & 14.6 & 14.0 & \textcolor{white}{0}8.0  & 12.4 & 20.3 & 16.8 & 19.4 & 19.3 \\
RMT ~\cite{dobler2023robust} & 24.1 & 20.2 & 25.7 & 13.2 & 25.5 & 14.7 & 12.8 & 16.2 & 15.4 & 14.6 & 10.8 & 14.0 & 18.0 & 14.1 & 16.6 & 17.0 \\
PETAL~\cite{brahma2023probabilistic}  & 23.7 & 21.4 & 26.3 & 11.8 & 28.8 & 12.4 & 10.4 & 14.8 & 13.9 & 12.6 & \textcolor{white}{0}\textbf{7.4}  & 10.6 & 18.3 & 13.1 & 17.1 & 16.2 \\
SATA~\cite{chakrabarty2023sata}   & 23.9 & 20.1 & 28.0 & 11.6 & 27.4 & 12.6 & 10.2 & 14.1 & 13.2 & 12.2 & \textcolor{white}{0}\textbf{7.4}  & 10.3 & 19.1 & 13.3 & 18.5 & 16.1 \\
DSS~\cite{wang2024continual}    & 24.1 & 21.3 & 25.4 & 11.7 & 26.9 & 12.2 & 10.5 & 14.5 & 14.1 & 12.5 & \textcolor{white}{0}7.8  & 10.8 & 18.0   & 13.1 & 17.3 & 16.0   \\
SWA~\cite{yang2023exploring}    & 23.9 & 20.5 & 24.5 & \textbf{11.2} & 26.3 & \textbf{11.8} & 10.1 & 14.0 & 12.7 & \textbf{11.5} & \textcolor{white}{0}7.6  & \textcolor{white}{0}\textbf{9.5}  & 17.6 & 12.0 & 15.8 & 15.3 \\
\midrule
Ours   & \textbf{22.7} & \textbf{17.9} & \textbf{23.8} & 11.6 & \textbf{24.3} & 12.8 & \textcolor{white}{0}\textbf{9.5}  & \textbf{13.1} & \textbf{12.4} & 11.6 & \textcolor{white}{0}8.0  & \textcolor{white}{0}\textbf{9.5}  & \textbf{16.4} & \textbf{11.4} & \textbf{15.6} & \textbf{14.7} \\
\bottomrule
\end{tabular}%
}
\vspace{-10px}

\end{table}

\textbf{Experiments on CIFAR100-C.}
To further demonstrate the effectiveness of the proposed method, we evaluate it on the more difficult CIFAR100-C task with the source-only baseline and eight SOTA methods. The experimental results are shown in Table~\ref{cifar100c}. Generally speaking, our method not only achieved the lowest error rates on the Snow, Frost, Fog, and Pix. tasks but also had the lowest average error rate. We improve the performance by 16.5\% and 0.4\% compared to the source-only baseline and SATA, respectively.
\begin{table}[t]
    \setlength{\belowcaptionskip}{-0.2cm}
    \caption{Classification error rate (\%) for standard CIFAR100-C continual test-time adaptation task.}
    \label{cifar100c}
    \centering
    \resizebox{\textwidth}{!}{%
\setlength\tabcolsep{3pt}
\begin{tabular}{@{}lcccccccccccccccc@{}}
\toprule
Method & Gauss. & Shot & Impul. & Defoc. & Glass & Motion & Zoom & Snow & Frost & Fog  & Brit. & Contr. & Elas. & Pix. & Jpeg & Mean \\ 
\midrule
Source & 73.0 & 68.0 & 39.4 & 29.3 & 54.1 & 30.8 & 28.8 & 39.5 & 45.8 & 50.3 & 29.5 & 55.1 & 37.2 & 74.7 & 41.2 & 46.4 \\
BN~\cite{li2017learning,schneider2020improving}     & 42.1 & 40.7 & 42.7 & 27.6 & 41.9 & 29.7 & 27.9 & 34.9 & 35.0 & 41.5 & 26.5 & 30.3 & 35.7 & 32.9 & 41.2 & 35.4 \\
TENT~\cite{wang2020tent}   & 37.2 & 35.8 & 41.7 & 37.9 & 51.2 & 48.3 & 48.5 & 58.4 & 63.7 & 71.1 & 70.4 & 82.3 & 88.0 & 88.5 & 90.4 & 60.9 \\
CoTTA~\cite{wang2022continual}  & 40.1 & 37.7 & 39.7 & 26.9 & 38.0 & 27.9 & 26.4 & 32.8 & 31.8 & 40.3 & 24.7 & 26.9 & 32.5 & 28.3 & 33.5 & 32.5 \\
RoTTA ~\cite{yuan2023robust} & 49.1 & 44.9 & 45.5 & 30.2 & 42.7 & 29.5 & 26.1 & 32.2 & 30.7 & 37.5 & 24.7 & 29.1 & 32.6 & 30.4 & 36.7 & 34.8 \\
RMT ~\cite{dobler2023robust} & 40.2 & 36.2 & 36.0 & 27.9 & \textbf{33.9} & 28.4 & 26.4 & \textbf{28.7} & 28.8 & 31.1 & 25.5 & 27.1 & \textbf{28.0} & 26.6 & \textbf{29.0} & 30.2 \\
PETAL~\cite{brahma2023probabilistic}  & 38.3 & 36.4 & 38.6 & \textbf{25.9} & 36.8 & \textbf{27.3} & 25.4 & 32.0 & 30.8 & 38.7 & 24.4 & 26.4 & 31.5 & 26.9 & 32.5 & 31.5 \\
DSS~\cite{wang2024continual}    & 39.7 & 36.0   & 37.2 & 26.3 & 35.6 & 27.5 & \textbf{25.1} & 31.4 & 30.0 & 37.8 & 24.2 & \textbf{26.0}  & 30.0  & 26.3 & 31.1 & 30.9 \\
SATA~\cite{chakrabarty2023sata}   & \textbf{36.5} & \textbf{33.1} & \textbf{35.1} & \textbf{25.9} & 34.9 & 27.7 & 25.4 & 29.5 & 29.9 & 33.1 & \textbf{23.6} & 26.7 & 31.9 & 27.5 & 35.2 & 30.3 \\
\midrule
Ours   & 38.1 & 34.8 & 36.4 & 27.1 & 34.3 & 27.7 & 26.1 & \textbf{28.7} & \textbf{28.5} & \textbf{30.9} & 24.1 & 26.2 & 28.2 & \textbf{26.2} & 31.2 & \textbf{29.9} \\
\bottomrule
\end{tabular}%
}
\vspace{-10px}

    \vspace{-10px}
\end{table}

\textbf{Experiments on ImageNet-C.}
The last experiment is conducted on ImageNet-C to further demonstrate the effectiveness of the proposed method. The experimental results can be seen in Table~\ref{imagenetc}. Compared with the SOTA methods, the proposed method achieve the lowest average error rate. Noteworthily, the proposed method outperforms SATA by a large margin for the Shot(71.6\% vs. 72.9\%), Impul.(68.7\% vs. 71.6\%), Defoc.(74.0\% vs. 75.7\%) and Glass.(71.6\% vs. 74.1\%) corruptions.
\begin{table}[t]
    \setlength{\belowcaptionskip}{-0.2cm}
    \caption{Classification error rate (\%) for standard ImageNet-C continual test-time adaptation task.}
    \label{imagenetc}
    \centering
    \resizebox{\textwidth}{!}{%
\setlength\tabcolsep{3pt}
\begin{tabular}{@{}lcccccccccccccccc@{}}
\toprule
Method & Gauss. & Shot & Impul. & Defoc. & Glass & Motion & Zoom & Snow & Frost & Fog  & Brit. & Contr. & Elas. & Pix. & Jpeg & Mean \\ 
\midrule
Source & 95.3 & 95.0 & 95.3 & 86.1 & 91.9 & 87.4 & 77.9 & 85.1 & 79.9 & 79.0 & 45.4 & 96.2 & 86.6 & 77.5 & 66.1 & 83.0 \\
BN~\cite{li2017learning,schneider2020improving}     & 87.7 & 87.4 & 87.8 & 88.0 & 87.7 & 78.3 & 63.9 & 67.4 & 70.3 & 54.7 & 36.4 & 88.7 & 58.0 & 56.6 & 67.0 & 72.0 \\
TENT~\cite{wang2020tent}   & 81.6 & 74.6 & 72.7 & 77.6 & 73.8 & 65.5 & \textbf{55.3} & 61.6 & 63.0 & 51.7 & 38.2 & 72.1 & 50.8 & 47.4 & 53.3 & 62.6 \\
CoTTA~\cite{wang2022continual}  & 84.7 & 82.1 & 80.6 & 81.3 & 79.0 & 68.6 & 57.5 & 60.3 & 60.5 & 48.3 & 36.6 & 66.1 & \textbf{47.2} & \textbf{41.2} & 46.0 & 62.7 \\
RoTTA ~\cite{yuan2023robust} & 88.3 & 82.8 & 82.1 & 91.3 & 83.7 & 72.9 & 59.4 & 66.2 & 64.3 & 53.3 & \textbf{35.6} & 74.5 & 54.3 & 48.2 & 52.6 & 67.3 \\
RMT ~\cite{dobler2023robust} & 79.9 & 76.3 & 73.1 & 75.7 & 72.9 & 64.7 & 56.8 & 56.4 & \textbf{58.3} & 49.0 & 40.6 & \textbf{58.2} & 47.8 & 43.7 & \textbf{44.8} & 59.9 \\
PETAL~\cite{brahma2023probabilistic}  & 87.4 & 85.8 & 84.4 & 85.0 & 83.9 & 74.4 & 63.1 & 63.5 & 64.0 & 52.4 & 40.0 & 74.0 & 51.7 & 45.2 & 51.0 & 67.1 \\
DSS~\cite{wang2024continual}    & 84.6 & 80.4 & 78.7 & 83.9 & 79.8 & 74.9 & 62.9 & 62.8 & 62.9 & 49.7 & 37.4 & 71.0   & 49.5 & 42.9 & 48.2 & 64.6 \\
ViDA~\cite{liu2023vida}   & 79.3 & 74.7 & 73.1 & 76.9 & 74.5 & 65.0 & 56.4 & 59.8 & 62.6 & 49.6 & 38.2 & 66.8 & 49.6 & 43.1 & 46.2 & 61.2 \\
SATA~\cite{chakrabarty2023sata}   & \textbf{74.1} & 72.9 & 71.6 & 75.7 & 74.1 & \textbf{64.2} & 55.5 & \textbf{55.6} & 62.9 & \textbf{46.6} & 36.1 & 69.9 & 50.6 & 44.3 & 48.5 & 60.1 \\
\midrule
Ours   & 75.1 & \textbf{71.6} & \textbf{68.7} & \textbf{74.0} & \textbf{71.6} & 65.1 & 56.4 & 55.7 & 61.0 & 49.3 & 41.3 & 61.9 & 49.1 & 44.8 & 46.0 & \textbf{59.4} \\
\bottomrule
\end{tabular}%
}


    \vspace{-10px}
\end{table}

\begin{table}[h]
    \centering
    \caption{Ablation study on class and domain under severity 5.}
    \vspace{-10px}
\resizebox{.65\linewidth}{!}{
\begin{tabular}{cccccc}
    \toprule
    No. & CCS & CDS & CIFAR10-C & CIFAR100-C & ImageNet-C \\
    \midrule
    1   &                   &               &   15.66    &      31.19     &     60.56          \\
    2   & \checkmark     &                  &   14.94    &      30.42     &     59.67          \\
    3   &       & \checkmark                &   15.16    &      30.68     &     59.75          \\
    4   & \checkmark     & \checkmark       &   14.71    &      29.92     &     59.43          \\
    \bottomrule
    \end{tabular}%
}
\label{AblationStudy}
    \vspace{-10px}
\end{table}
\subsection{Ablation Studies}
We perform ablation study experiments to evaluate the effectiveness of major components of C-CoTTA on three benchmarks. For simplicity, we denote the Control Class Shift as CCS and Control Domain Shift as CDS. As shown in Table~\ref{AblationStudy}, C-CoTTA decrease error rates in all benchmarks after adding CCS or CDS to the network, which indicate mitigating inter-category interference or reducing the model's sensitivity to overall domain shifts can improve the classification accuracy. Furthermore, the combination of CCS and CDS can further improve the classification accuracy of the C-CoTTA framework.

\subsection{t-SNE Visualization for Class Shift}
\begin{figure}[t]
    \centering
    \includegraphics[scale=0.275]{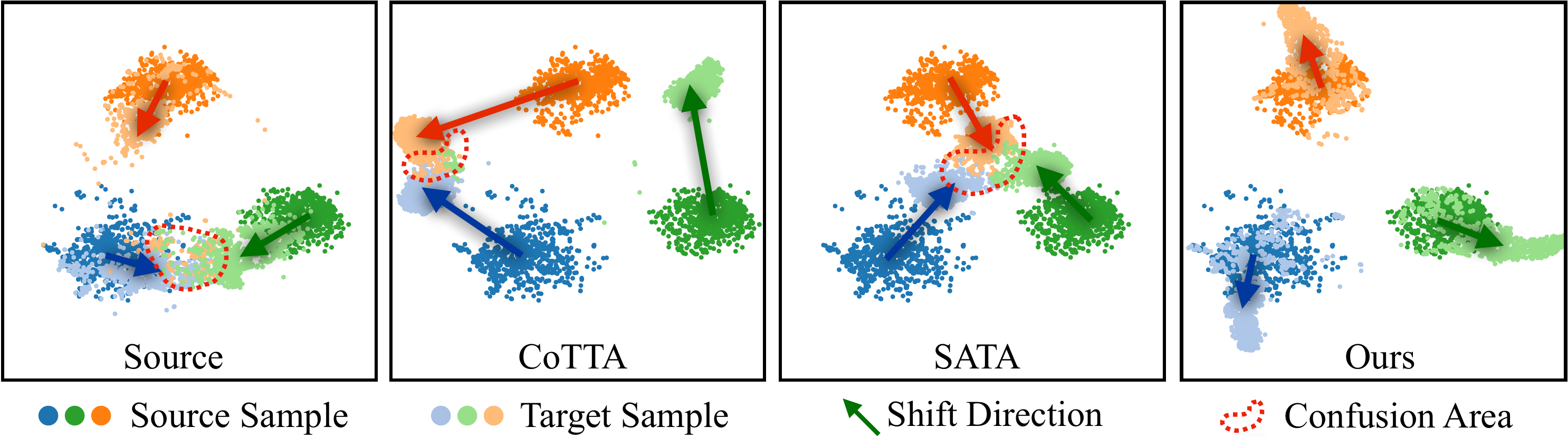}
    \caption{Visualization of the t-SNE dimensionality reduction of three classes from CIFAR10-C dataset (three easily misclassified animals: bird, deer, frog) transferred from the source domain to the target domain (zoom)}
    \label{fig:tsne_mul_mt}
    \vspace{-10px}
\end{figure}

We use t-SNE~\cite{van2008visualizing} for dimensionality reduction to visualize domain shift situations of different methods during test-time. As shown in Figure~\ref{fig:tsne_mul_mt}, it can be observed that directly using pre-trained model(Source), CoTTA, and SATA methods do not explicitly control domain shift, resulting in unpredictable domain shift directions, leading to blurry classification boundaries and mutual interference between categories. In contrast, our method implements controllable domain shift, so it can be seen that although categories also experience shift, the direction of the shift is benign and does not cause mutual interference between categories.
%

\subsection{Inter-Class Distance and Inter-Domain Distance}
\begin{figure}[h!]
    \setlength{\abovecaptionskip}{0pt}
    \centering
    \includegraphics[scale=0.22]{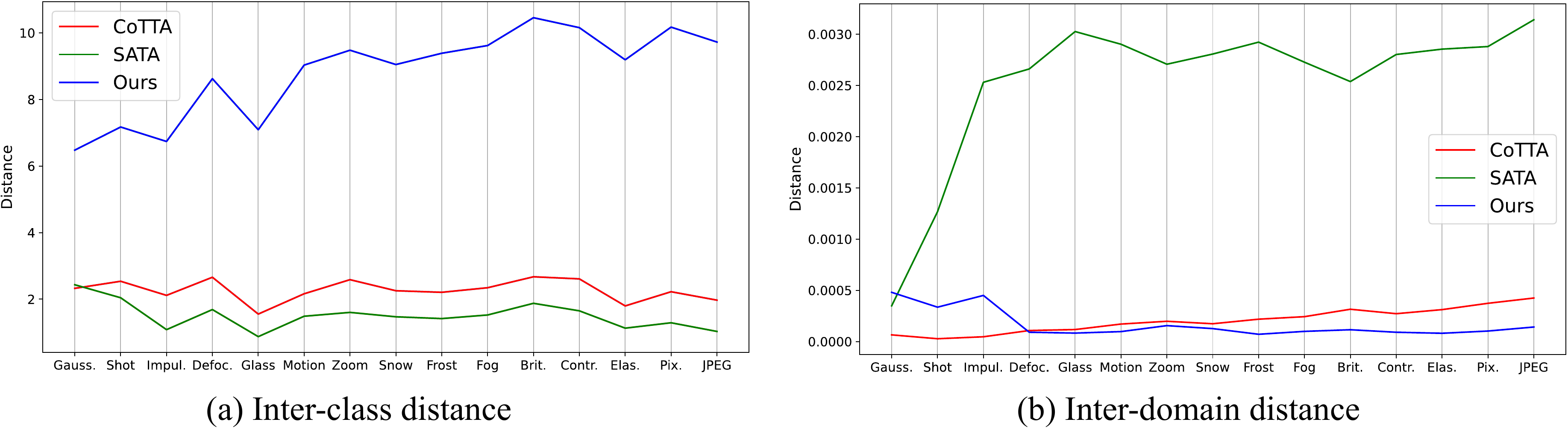}
    \caption{(a) Inter-class distance: can indicate the separability between classes. (b) Inter-domain distance: can indicate the sensitivity of the model to domain transformations.}
    \label{fig:distence}
\end{figure}

\textbf{Inter-Class Distance.} The inter-class distance can be formulate as $d_{ic}=\sum_{i=1}^{c}  \sum_{j \ne i}^{c} \left\|\mathbf{p}_{i}^t-\mathbf{p}_{j}^t\right\|_{2}^{2}$, where $\mathbf{p}_{i}^t$ and $\mathbf{p}_{j}^t$ denote target domain category prototype.
The inter-class distance comparison between our method, CoTTA and SATA can be seen in Figure~\ref{fig:distence}(a). Compared to CoTTA and SATA, our method has a larger inter-class distance, indicating better separability between classes and reflecting the effectiveness of controlling class shift direction.

\textbf{Inter-Domain Distance.} The inter-domain distance can be computed as $d_{id}=\left\|\mathbf{p}^{s}-\mathbf{p}^{t}\right\|_{2}^{2}$, where $\mathbf{p}^{s}$ denotes overall prototype of the source domain while $\mathbf{p}^{t}$ denotes overall prototype of the target domain. As shown in the Figure~\ref{fig:distence}(b), compared to SATA, our method has smaller inter-domain distances, indicating that the model is less sensitive to domain transformations and reflecting the effectiveness of controlling domain shift. At the same time, although the CoTTA method has relatively small inter-domain distances in the early stage, as the target domain changes at test time, its inter-domain distances gradually increase. This indicates that the model is sensitive to domain shift, further emphasizing the importance of reducing the sensitivity of the model to domain shift in controlling overall domain shift.

%

\subsection{Results for Gradual Test-Time Adaptation}
In the standard setup described above, corruption types change abruptly at the highest severity level. We will now present the results of the gradual setup. We design the sequence by gradually changing the severity for the 15 types of corruption. When the type of corruption changes, the severity level is at its lowest.  The distribution shift within each type is also gradual. Table~\ref{tab:gradually} shows that the proposed method outperforms competing methods.

\begin{figure}[t]

\begin{minipage}[t]{0.5\textwidth}
\makeatletter\def\@captype{table}
\caption{Results for Gradual Adaptation}
\resizebox{!}{1.18cm}{
\begin{tabular}{cccc}
    \toprule
    Methods & CIFAR10-C & CIFAR100-C & ImageNet-C \\
    \midrule
    Source & 23.9     & 32.9      & 81.7      \\
    TENT   & 39.1     & 72.7      & 53.7      \\
    CoTTA  & 10.6     & 26.3      & 42.1      \\
    SATA    &  10.8   &  27.5      & 44.8         \\
    Ours   & \textbf{9.21}  & \textbf{26.2}         & \textbf{41.1}         \\
    \bottomrule
    \end{tabular}%
}

\label{tab:gradually}
\end{minipage}
\begin{minipage}[t]{0.5\textwidth}
\makeatletter\def\@captype{table}
\caption{Results for Corruption loops Adaptation}
\resizebox{!}{1.18cm}{
\begin{tabular}{cccc}
    \toprule
    Methods & CIFAR10-C & CIFAR100-C & ImageNet-C \\
    \midrule
    Source  &    43.5      &        46.4      &         83.0      \\
    TENT    &    41.8      &        31.2         &      65.3       \\
    CoTTA   &    15.7      &        32.4         &       68.2        \\
    SATA    &    15.5      &        32.2        &        62.5           \\
    Ours    &    \textbf{11.6}      &       \textbf{27.2}           &          \textbf{52.6}         \\
    \bottomrule
    \end{tabular}%
}
\label{tab:loop}
\end{minipage}
\vspace{-10px}
\end{figure}

\subsection{Results for Corruption Loops Test-Time Adaptation}

In real-world scenarios, test domains may occur in cycles. To assess the long-term adaptation performance of the method, we evaluated the test conditions for 10 consecutive cycles. This means that at level 5 severity, the test data will be reanalyzed and readjusted nine times. The complete result can be found in Table~\ref{tab:loop}. The results also show that our method outperforms the other methods in this long-term adaptation scenario.  It illustrates the effectiveness of domain offset controllability as well as category offset controllability.

\subsection{Results for Random Order Test-Time Adaptation}

For a more comprehensive evaluation of the proposed method, CIFAR10-C, CIFAR100-C, and ImageNet-C experiments are conducted on over ten sequences of various damage types with a severity level of 5. As shown in the Table~\ref{tab:random}, C-CoTTA is consistently outperforming CoTTA and other competing methods.

\begin{table}[h]
    \centering
    \caption{Average error of standard ImageNet-C experiments over 10 diverse corruption sequences.}
    \vspace{-10px}
\resizebox{.6\linewidth}{!}{
\begin{tabular}{cccccc}
\toprule
Avg. Error (\%) & Source & TENT  & CoTTA  & SATA & Ours          \\
\midrule
CIFAR10-C   & 43.5     & 20.1   & 16.3      &  16.3     &  \textbf{14.7}     \\
CIFAR100-C  & 46.4     & 61.3   & 32.6      &  32.8     &  \textbf{29.5}     \\
ImageNet-C  & 83.0     & 61.8   & 57.9      &  64.5     &  \textbf{55.5}     \\
\bottomrule
\end{tabular}
}

\label{tab:random}
\vspace{-10px}
\end{table}

\section{Conclusion and Limitation}

In this work, we introduce C-CoTTA, a novel framework designed to prevent any category from leaning towards other categories by explicitly controlling the offset direction to avoid fuzzy classification boundaries, and reduce the sensitivity of the model in the domain shift direction to reduce the impact of domain shift on domain adaptation. This fills the gap left by traditional methods, which can only mitigate the impact of domain drift. C-CoTTA can explicitly control domain shift, opening up a new solution pathway for CTTA. Through extensive quantitative experiments and qualitative analysis, such as t-SNE plots, we demonstrate the effectiveness and theoretical validity of C-CoTTA.

However, our method also has certain limitations. During test-time, we may not have access to all samples of a specific domain prototype, and due to the lack of true labels, misclassified samples may contaminate the prototype. As a result, the representation of the prototype may be poor, further affecting the accuracy of the constructed direction, leading to ineffective or even erroneous domain shift control. Therefore, in the future, addressing how to obtain high-quality prototypes and directional representations is a task that needs attention.

\medskip

{
\small
\bibliographystyle{plainnat}
\bibliography{cite}

\begin{thebibliography}{54}
\providecommand{\natexlab}[1]{#1}
\providecommand{\url}[1]{\texttt{#1}}
\expandafter\ifx\csname urlstyle\endcsname\relax
  \providecommand{\doi}[1]{doi: #1}\else
  \providecommand{\doi}{doi: \begingroup \urlstyle{rm}\Url}\fi

\bibitem[Anders et~al.(2022)Anders, Weber, Neumann, Samek, M{\"u}ller, and
  Lapuschkin]{anders2022finding}
Christopher~J Anders, Leander Weber, David Neumann, Wojciech Samek,
  Klaus-Robert M{\"u}ller, and Sebastian Lapuschkin.
\newblock Finding and removing clever hans: Using explanation methods to debug
  and improve deep models.
\newblock \emph{Information Fusion}, 77:\penalty0 261--295, 2022.

\bibitem[Biecek and Samek(2024)]{biecek2024explain}
Przemyslaw Biecek and Wojciech Samek.
\newblock Explain to question not to justify.
\newblock \emph{arXiv preprint arXiv:2402.13914}, 2024.

\bibitem[Brahma and Rai(2023)]{brahma2023probabilistic}
Dhanajit Brahma and Piyush Rai.
\newblock A probabilistic framework for lifelong test-time adaptation.
\newblock In \emph{Proceedings of the IEEE/CVF Conference on Computer Vision
  and Pattern Recognition}, pages 3582--3591, 2023.

\bibitem[Chakrabarty et~al.(2023)Chakrabarty, Sreenivas, and
  Biswas]{chakrabarty2023sata}
Goirik Chakrabarty, Manogna Sreenivas, and Soma Biswas.
\newblock Sata: Source anchoring and target alignment network for continual
  test time adaptation.
\newblock \emph{arXiv preprint arXiv:2304.10113}, 2023.

\bibitem[Chen(2020)]{chen2020domain}
Xingxin Chen.
\newblock Domain adaptation for autonomous driving.
\newblock Master's thesis, University of Waterloo, 2020.

\bibitem[Chen et~al.(2023)Chen, Ye, Lu, Pan, and Xia]{chen2023each}
Ziyang Chen, Yiwen Ye, Mengkang Lu, Yongsheng Pan, and Yong Xia.
\newblock Each test image deserves a specific prompt: Continual test-time
  adaptation for 2d medical image segmentation.
\newblock \emph{arXiv preprint arXiv:2311.18363}, 2023.

\bibitem[Cortes and Vapnik(1995)]{cortes1995support}
Corinna Cortes and Vladimir Vapnik.
\newblock Support-vector networks.
\newblock \emph{Machine learning}, 20:\penalty0 273--297, 1995.

\bibitem[D{\"o}bler et~al.(2023)D{\"o}bler, Marsden, and
  Yang]{dobler2023robust}
Mario D{\"o}bler, Robert~A Marsden, and Bin Yang.
\newblock Robust mean teacher for continual and gradual test-time adaptation.
\newblock In \emph{Proceedings of the IEEE/CVF Conference on Computer Vision
  and Pattern Recognition}, pages 7704--7714, 2023.

\bibitem[Dreyer et~al.(2023)Dreyer, Pahde, Anders, Samek, and
  Lapuschkin]{dreyer2023hope}
Maximilian Dreyer, Frederik Pahde, Christopher~J Anders, Wojciech Samek, and
  Sebastian Lapuschkin.
\newblock From hope to safety: Unlearning biases of deep models by enforcing
  the right reasons in latent space.
\newblock \emph{arXiv preprint arXiv:2308.09437}, 2023.

\bibitem[Dreyer et~al.(2024)Dreyer, Pahde, Anders, Samek, and
  Lapuschkin]{dreyer2024hope}
Maximilian Dreyer, Frederik Pahde, Christopher~J Anders, Wojciech Samek, and
  Sebastian Lapuschkin.
\newblock From hope to safety: Unlearning biases of deep models via gradient
  penalization in latent space.
\newblock In \emph{Proceedings of the AAAI Conference on Artificial
  Intelligence}, volume~38, pages 21046--21054, 2024.

\bibitem[Du et~al.(2023)Du, Lyu, Li, Hu, Feng, Xu, Xi, and Cheng]{du2023multi}
Kaile Du, Fan Lyu, Linyan Li, Fuyuan Hu, Wei Feng, Fenglei Xu, Xuefeng Xi, and
  Hanjing Cheng.
\newblock Multi-label continual learning using augmented graph convolutional
  network.
\newblock \emph{IEEE Transactions on Multimedia}, 2023.

\bibitem[Gan et~al.(2023)Gan, Bai, Lou, Ma, Zhang, Shi, and
  Luo]{gan2023decorate}
Yulu Gan, Yan Bai, Yihang Lou, Xianzheng Ma, Renrui Zhang, Nian Shi, and Lin
  Luo.
\newblock Decorate the newcomers: Visual domain prompt for continual test time
  adaptation.
\newblock In \emph{Proceedings of the AAAI Conference on Artificial
  Intelligence}, volume~37, pages 7595--7603, 2023.

\bibitem[Geiger et~al.(2013)Geiger, Lenz, Stiller, and
  Urtasun]{geiger2013vision}
Andreas Geiger, Philip Lenz, Christoph Stiller, and Raquel Urtasun.
\newblock Vision meets robotics: The kitti dataset.
\newblock \emph{The International Journal of Robotics Research}, 32\penalty0
  (11):\penalty0 1231--1237, 2013.

\bibitem[Gonzalez et~al.(2020)Gonzalez, Lemke, Sakas, and
  Mukhopadhyay]{gonzalez2020wrong}
Camila Gonzalez, Nick Lemke, Georgios Sakas, and Anirban Mukhopadhyay.
\newblock What is wrong with continual learning in medical image segmentation?
\newblock \emph{arXiv preprint arXiv:2010.11008}, 2020.

\bibitem[Haufe et~al.(2014)Haufe, Meinecke, G{\"o}rgen, D{\"a}hne, Haynes,
  Blankertz, and Bie{\ss}mann]{haufe2014interpretation}
Stefan Haufe, Frank Meinecke, Kai G{\"o}rgen, Sven D{\"a}hne, John-Dylan
  Haynes, Benjamin Blankertz, and Felix Bie{\ss}mann.
\newblock On the interpretation of weight vectors of linear models in
  multivariate neuroimaging.
\newblock \emph{Neuroimage}, 87:\penalty0 96--110, 2014.

\bibitem[He et~al.(2016)He, Zhang, Ren, and Sun]{he2016deep}
Kaiming He, Xiangyu Zhang, Shaoqing Ren, and Jian Sun.
\newblock Deep residual learning for image recognition.
\newblock In \emph{Proceedings of the Computer Vision and Pattern Recognition},
  2016.

\bibitem[Hu et~al.(2022)Hu, Hudson, Ethier, Al-Sharman, Rayside, and
  Melek]{hu2022sim}
Chuqing Hu, Sinclair Hudson, Martin Ethier, Mohammad Al-Sharman, Derek Rayside,
  and William Melek.
\newblock Sim-to-real domain adaptation for lane detection and classification
  in autonomous driving.
\newblock In \emph{IEEE Intelligent Vehicles Symposium (IV)}, pages 457--463.
  IEEE, 2022.

\bibitem[Jain and Learned-Miller(2011)]{jain2011online}
Vidit Jain and Erik Learned-Miller.
\newblock Online domain adaptation of a pre-trained cascade of classifiers.
\newblock In \emph{CVPR}, pages 577--584. IEEE, 2011.

\bibitem[Kim et~al.(2018)Kim, Wattenberg, Gilmer, Cai, Wexler, Viegas,
  et~al.]{kim2018interpretability}
Been Kim, Martin Wattenberg, Justin Gilmer, Carrie Cai, James Wexler, Fernanda
  Viegas, et~al.
\newblock Interpretability beyond feature attribution: Quantitative testing
  with concept activation vectors (tcav).
\newblock In \emph{International conference on machine learning}, pages
  2668--2677. PMLR, 2018.

\bibitem[Kondo(2022)]{kondo2022source}
Satoshi Kondo.
\newblock Source-free unsupervised domain adaptation with norm and shape
  constraints for medical image segmentation.
\newblock \emph{arXiv preprint arXiv:2209.01300}, 2022.

\bibitem[Li and Hoiem(2017)]{li2017learning}
Zhizhong Li and Derek Hoiem.
\newblock Learning without forgetting.
\newblock \emph{IEEE transactions on pattern analysis and machine
  intelligence}, 40\penalty0 (12):\penalty0 2935--2947, 2017.

\bibitem[Liu et~al.(2023)Liu, Yang, Jia, Lu, Guo, Xue, and Zhang]{liu2023vida}
Jiaming Liu, Senqiao Yang, Peidong Jia, Ming Lu, Yandong Guo, Wei Xue, and
  Shanghang Zhang.
\newblock Vida: Homeostatic visual domain adapter for continual test time
  adaptation.
\newblock \emph{arXiv preprint arXiv:2306.04344}, 2023.

\bibitem[Lyu et~al.(2021)Lyu, Wang, Feng, Ye, Hu, and Wang]{lyu2021multi}
Fan Lyu, Shuai Wang, Wei Feng, Zihan Ye, Fuyuan Hu, and Song Wang.
\newblock Multi-domain multi-task rehearsal for lifelong learning.
\newblock In \emph{Proceedings of the AAAI Conference on Artificial
  Intelligence}, volume~35, pages 8819--8827, 2021.

\bibitem[Lyu et~al.(2024{\natexlab{a}})Lyu, Du, Li, Zhao, Zhang, Liu, and
  Wang]{lyu2024variational}
Fan Lyu, Kaile Du, Yuyang Li, Hanyu Zhao, Zhang Zhang, Guangcan Liu, and Liang
  Wang.
\newblock Variational continual test-time adaptation.
\newblock \emph{arXiv preprint arXiv:2402.08182}, 2024{\natexlab{a}}.

\bibitem[Lyu et~al.(2024{\natexlab{b}})Lyu, Feng, Li, Sun, Shang, Wan, and
  Wang]{lyu2024elastic}
Fan Lyu, Wei Feng, Yuepan Li, Qing Sun, Fanhua Shang, Liang Wan, and Liang
  Wang.
\newblock Elastic multi-gradient descent for parallel continual learning.
\newblock \emph{arXiv preprint arXiv:2401.01054}, 2024{\natexlab{b}}.

\bibitem[McGrath et~al.(2022)McGrath, Kapishnikov, Toma{\v{s}}ev, Pearce,
  Wattenberg, Hassabis, Kim, Paquet, and Kramnik]{mcgrath2022acquisition}
Thomas McGrath, Andrei Kapishnikov, Nenad Toma{\v{s}}ev, Adam Pearce, Martin
  Wattenberg, Demis Hassabis, Been Kim, Ulrich Paquet, and Vladimir Kramnik.
\newblock Acquisition of chess knowledge in alphazero.
\newblock \emph{Proceedings of the National Academy of Sciences}, 119\penalty0
  (47):\penalty0 e2206625119, 2022.

\bibitem[Ni et~al.(2023)Ni, Yang, Liu, Li, Jiao, Xu, Chen, Liu, and
  Zhang]{ni2023distribution}
Jiayi Ni, Senqiao Yang, Jiaming Liu, Xiaoqi Li, Wenyu Jiao, Ran Xu, Zehui Chen,
  Yi~Liu, and Shanghang Zhang.
\newblock Distribution-aware continual test time adaptation for semantic
  segmentation.
\newblock \emph{arXiv preprint arXiv:2309.13604}, 2023.

\bibitem[Niloy et~al.(2024)Niloy, Ahmed, Raychaudhuri, Oymak, and
  Roy-Chowdhury]{niloy2024effective}
Fahim~Faisal Niloy, Sk~Miraj Ahmed, Dripta~S Raychaudhuri, Samet Oymak, and
  Amit~K Roy-Chowdhury.
\newblock Effective restoration of source knowledge in continual test time
  adaptation.
\newblock In \emph{Proceedings of the IEEE/CVF Winter Conference on
  Applications of Computer Vision}, pages 2091--2100, 2024.

\bibitem[Niu et~al.(2022)Niu, Wu, Zhang, Chen, Zheng, Zhao, and
  Tan]{niu2022eata}
Shuaicheng Niu, Jiaxiang Wu, Yifan Zhang, Yaofo Chen, Shijian Zheng, Peilin
  Zhao, and Mingkui Tan.
\newblock Efficient test-time model adaptation without forgetting, 2022.

\bibitem[O'Kelly(2021)]{o2021accelerated}
Matthew O'Kelly.
\newblock \emph{Accelerated Risk Assessment and Domain Adaptation for
  Autonomous Vehicles}.
\newblock PhD thesis, University of Pennsylvania, 2021.

\bibitem[Pahde et~al.(2022)Pahde, Weber, Anders, Samek, and
  Lapuschkin]{pahde2022patclarc}
Frederik Pahde, Leander Weber, Christopher~J Anders, Wojciech Samek, and
  Sebastian Lapuschkin.
\newblock Patclarc: Using pattern concept activation vectors for noise-robust
  model debugging.
\newblock \emph{arXiv preprint arXiv:2202.03482}, 2022.

\bibitem[Pahde et~al.(2023)Pahde, Dreyer, Samek, and
  Lapuschkin]{pahde2023reveal}
Frederik Pahde, Maximilian Dreyer, Wojciech Samek, and Sebastian Lapuschkin.
\newblock Reveal to revise: An explainable ai life cycle for iterative bias
  correction of deep models.
\newblock In \emph{Medical Image Computing and Computer Assisted Intervention},
  2023.

\bibitem[Pfau et~al.(2021)Pfau, Young, Wei, Wei, and Keiser]{pfau2021robust}
Jacob Pfau, Albert~T Young, Jerome Wei, Maria~L Wei, and Michael~J Keiser.
\newblock Robust semantic interpretability: Revisiting concept activation
  vectors.
\newblock \emph{arXiv preprint arXiv:2104.02768}, 2021.

\bibitem[Rahman et~al.(2021)Rahman, Fookes, and Sridharan]{rahman2021deep}
Mohammad~Mahfujur Rahman, Clinton Fookes, and Sridha Sridharan.
\newblock Deep domain generalization with feature-norm network.
\newblock \emph{arXiv preprint arXiv:2104.13581}, 2021.

\bibitem[Samek(2023)]{samek2023explainable}
Wojciech Samek.
\newblock Explainable deep learning: concepts, methods, and new developments.
\newblock In \emph{Explainable Deep Learning AI}, pages 7--33. Elsevier, 2023.

\bibitem[Schneider et~al.(2020)Schneider, Rusak, Eck, Bringmann, Brendel, and
  Bethge]{schneider2020improving}
Steffen Schneider, Evgenia Rusak, Luisa Eck, Oliver Bringmann, Wieland Brendel,
  and Matthias Bethge.
\newblock Improving robustness against common corruptions by covariate shift
  adaptation.
\newblock In \emph{Proceedings of the Advances in Neural Information Processing
  Systems}, 2020.

\bibitem[Sohn et~al.(2020)Sohn, Berthelot, Carlini, Zhang, Zhang, Raffel,
  Cubuk, Kurakin, and Li]{sohn2020fixmatch}
Kihyuk Sohn, David Berthelot, Nicholas Carlini, Zizhao Zhang, Han Zhang,
  Colin~A Raffel, Ekin~Dogus Cubuk, Alexey Kurakin, and Chun-Liang Li.
\newblock Fixmatch: Simplifying semi-supervised learning with consistency and
  confidence.
\newblock \emph{Advances in neural information processing systems},
  33:\penalty0 596--608, 2020.

\bibitem[Sun et~al.(2022)Sun, Lyu, Shang, Feng, and Wan]{sun2022exploring}
Qing Sun, Fan Lyu, Fanhua Shang, Wei Feng, and Liang Wan.
\newblock Exploring example influence in continual learning.
\newblock In \emph{NeurIPS 2022 (Oral)}, 2022.

\bibitem[Sun et~al.(2020)Sun, Wang, Liu, Miller, Efros, and Hardt]{sun2020test}
Yu~Sun, Xiaolong Wang, Zhuang Liu, John Miller, Alexei Efros, and Moritz Hardt.
\newblock Test-time training with self-supervision for generalization under
  distribution shifts.
\newblock In \emph{International conference on machine learning}, pages
  9229--9248. PMLR, 2020.

\bibitem[Tarvainen and Valpola(2017)]{tarvainen2017mean}
Antti Tarvainen and Harri Valpola.
\newblock Mean teachers are better role models: Weight-averaged consistency
  targets improve semi-supervised deep learning results.
\newblock \emph{Advances in neural information processing systems}, 30, 2017.

\bibitem[Van~de Ven and Tolias(2019)]{van2019three}
Gido~M Van~de Ven and Andreas~S Tolias.
\newblock Three scenarios for continual learning.
\newblock \emph{arXiv preprint arXiv:1904.07734}, 2019.

\bibitem[Van~der Maaten and Hinton(2008)]{van2008visualizing}
Laurens Van~der Maaten and Geoffrey Hinton.
\newblock Visualizing data using t-sne.
\newblock \emph{Journal of machine learning research}, 9\penalty0 (11), 2008.

\bibitem[Wang et~al.(2020)Wang, Shelhamer, Liu, Olshausen, and
  Darrell]{wang2020tent}
Dequan Wang, Evan Shelhamer, Shaoteng Liu, Bruno Olshausen, and Trevor Darrell.
\newblock Tent: Fully test-time adaptation by entropy minimization.
\newblock \emph{arXiv preprint arXiv:2006.10726}, 2020.

\bibitem[Wang et~al.(2022)Wang, Fink, Van~Gool, and Dai]{wang2022continual}
Qin Wang, Olga Fink, Luc Van~Gool, and Dengxin Dai.
\newblock Continual test-time domain adaptation.
\newblock In \emph{Proceedings of the IEEE/CVF Conference on Computer Vision
  and Pattern Recognition}, pages 7201--7211, 2022.

\bibitem[Wang et~al.(2024)Wang, Hong, Cheraghian, Rahman, Ahmedt-Aristizabal,
  Petersson, and Harandi]{wang2024continual}
Yanshuo Wang, Jie Hong, Ali Cheraghian, Shafin Rahman, David
  Ahmedt-Aristizabal, Lars Petersson, and Mehrtash Harandi.
\newblock Continual test-time domain adaptation via dynamic sample selection.
\newblock In \emph{Proceedings of the IEEE/CVF Winter Conference on
  Applications of Computer Vision}, pages 1701--1710, 2024.

\bibitem[Wang et~al.(2019)Wang, Ma, Chen, Luo, Yi, and
  Bailey]{wang2019symmetric}
Yisen Wang, Xingjun Ma, Zaiyi Chen, Yuan Luo, Jinfeng Yi, and James Bailey.
\newblock Symmetric cross entropy for robust learning with noisy labels.
\newblock In \emph{Proceedings of the IEEE/CVF International Conference on
  Computer Vision}, pages 322--330, 2019.

\bibitem[Weber et~al.(2023)Weber, Lapuschkin, Binder, and
  Samek]{weber2023beyond}
Leander Weber, Sebastian Lapuschkin, Alexander Binder, and Wojciech Samek.
\newblock Beyond explaining: Opportunities and challenges of xai-based model
  improvement.
\newblock \emph{Information Fusion}, 92:\penalty0 154--176, 2023.

\bibitem[Xie et~al.(2020)Xie, Luong, Hovy, and Le]{xie2020self}
Qizhe Xie, Minh-Thang Luong, Eduard Hovy, and Quoc~V Le.
\newblock Self-training with noisy student improves imagenet classification.
\newblock In \emph{Proceedings of the IEEE/CVF conference on computer vision
  and pattern recognition}, pages 10687--10698, 2020.

\bibitem[Xie et~al.(2017)Xie, Girshick, Doll{\'a}r, Tu, and
  He]{xie2017aggregated}
Saining Xie, Ross Girshick, Piotr Doll{\'a}r, Zhuowen Tu, and Kaiming He.
\newblock Aggregated residual transformations for deep neural networks.
\newblock In \emph{Proceedings of the Computer Vision and Pattern Recognition},
  2017.

\bibitem[Xu et~al.(2019)Xu, Li, Yang, and Lin]{xu2019larger}
Ruijia Xu, Guanbin Li, Jihan Yang, and Liang Lin.
\newblock Larger norm more transferable: An adaptive feature norm approach for
  unsupervised domain adaptation.
\newblock In \emph{Proceedings of the IEEE/CVF international conference on
  computer vision}, pages 1426--1435, 2019.

\bibitem[Yang et~al.(2023)Yang, Gu, Wei, and Deng]{yang2023exploring}
Xu~Yang, Yanan Gu, Kun Wei, and Cheng Deng.
\newblock Exploring safety supervision for continual test-time domain
  adaptation.
\newblock In \emph{Proceedings of the Thirty-Second International Joint
  Conference on Artificial Intelligence}, pages 1649--1657, 2023.

\bibitem[Yuan et~al.(2023)Yuan, Xie, and Li]{yuan2023robust}
Longhui Yuan, Binhui Xie, and Shuang Li.
\newblock Robust test-time adaptation in dynamic scenarios.
\newblock In \emph{Proceedings of the IEEE/CVF Conference on Computer Vision
  and Pattern Recognition}, pages 15922--15932, 2023.

\bibitem[Yuksekgonul et~al.(2022)Yuksekgonul, Wang, and
  Zou]{yuksekgonul2022post}
Mert Yuksekgonul, Maggie Wang, and James Zou.
\newblock Post-hoc concept bottleneck models.
\newblock \emph{arXiv preprint arXiv:2205.15480}, 2022.

\bibitem[Zagoruyko and Komodakis(2016)]{zagoruyko2016wide}
Sergey Zagoruyko and Nikos Komodakis.
\newblock Wide residual networks.
\newblock In \emph{Procedings of the British Machine Vision Conference}, 2016.

\end{thebibliography}
}

\newpage
\appendix

\section{Equivalent Representation of SCAV}
\label{sec:appendix1}

Further derivation of scav is conducted to obtain a simpler equivalent representation, as follows:

\begin{align*}
\text{cov}[f(x),y] &=\sum_{x_{i}\in \mathcal{X}_{c}\cup \mathcal{X}_{n}}(f(x_{i})-\overline{f}(x))(y_{i}-\overline{y}) \\
&=\frac{|\mathcal{X}_{n}|}{|\mathcal{X}_{c}|+|\mathcal{X}_{n}|}\sum_{x_{i}\in \mathcal{X}_{c}}(f(x_{i})-\overline{f}(x))+\frac{|\mathcal{X}_{c}|}{|\mathcal{X}_{c}|+|\mathcal{X}_{n}|}\sum_{x_{i}\in \mathcal{X}_{n}}(f(x_{i})-\overline{f}(x)) \\
&=\frac{|\mathcal{X}_{n}|}{|\mathcal{X}_{c}|+|\mathcal{X}_{n}|}\left(\sum_{x_{i}\in \mathcal{X}_{c}}f(x_{i})-|\mathcal{X}_{c}|\cdot\overline{f}(x_{i})\right)-\frac{|\mathcal{X}_{c}|}{|\mathcal{X}_{c}|+|\mathcal{X}_{n}|}\left(\sum_{x_{i}\in \mathcal{X}_{n}}f(x_{i})-|\mathcal{X}_{n}|\overline{f}(x_{i})\right) \\
&=\frac{|\mathcal{X}_{n}|}{|\mathcal{X}_{c}|+|\mathcal{X}_{n}|}\sum_{x_{i}\in \mathcal{X}_{c}}f(x_{i})-\frac{|\mathcal{X}_{n}||\mathcal{X}_{c}|}{|\mathcal{X}_{c}|+|\mathcal{X}_{n}|}\overline{f}(x_{i})-\frac{|\mathcal{X}_{c}|}{|\mathcal{X}_{c}|+|\mathcal{X}_{n}|}\sum_{x_{i}\in \mathcal{X}_{n}}f(x_{i})+\frac{|\mathcal{X}_{c}||\mathcal{X}_{n}|}{|\mathcal{X}_{c}|+|\mathcal{X}_{n}|}\overline{f}(x_{i}) \\
&=\frac{|\mathcal{X}_{n}||\mathcal{X}_{c}|}{|\mathcal{X}_{c}|+|\mathcal{X}_{n}|}\left(\frac{1}{|\mathcal{X}_{c}|}{\sum_{x_{i}\in \mathcal{X}_{c}}f(x_{i})}-\frac{1}{|\mathcal{X}_{n}|}{\sum_{x_{i}\in \mathcal{X}_{n}}f(x_{i})}\right) \\
\text{cov}[y,y] &=\sum_{x_{i}\in \mathcal{X}_{c} \cup \mathcal{X}_{n}}(y_{i}-\overline{y})^{2} \\
&=\sum_{x_{i}\in \mathcal{X}_{c}}(1-\frac{|\mathcal{X}_{c}|}{|\mathcal{X}_{c}|+|\mathcal{X}_{n}|})^{2}+\sum_{x_{i}\in \mathcal{X}_{n}}(0-\frac{|\mathcal{X}_{c}|}{|\mathcal{X}_{c}|+|\mathcal{X}_{n}|})^{2} \\
&=\sum_{x_{i}\in \mathcal{X}_{c}}\frac{|\mathcal{X}_{n}|^{2}}{(|\mathcal{X}_{c}|+|\mathcal{X}_{n}|)^{2}}+\sum_{x_{i}\in \mathcal{X}_{n}}\frac{|\mathcal{X}_{c}|^{2}}{(|\mathcal{X}_{c}|+|\mathcal{X}_{n}|)^{2}} \\
&=\frac{|\mathcal{X}_{c}||\mathcal{X}_{n}|^{2}+|\mathcal{X}_{n}||\mathcal{X}_{c}|^{2}}{(|\mathcal{X}_{c}+|\mathcal{X}_{n}|)^{2}} \\
&=\frac{|\mathcal{X}_{c}||\mathcal{X}_{n}|}{|\mathcal{X}_{c}|+|X_{n}|} \\
\frac{\text{cov}[f(x),y]}{\text{cov}[y,y]} &=\frac{1}{|\mathcal{X}_{c}|}{\sum_{x_{i}\in \mathcal{X}_{c}}f(x_{i})}-\frac{1}{|\mathcal{X}_{n}|}{\sum_{x_{i}\in \mathcal{X}_{n}}f(x_{i})}
\end{align*}

\section{Corruption loops Test-Time Adaptation}

\begin{figure}[h!]
    \centering
    \includegraphics[scale=0.21]{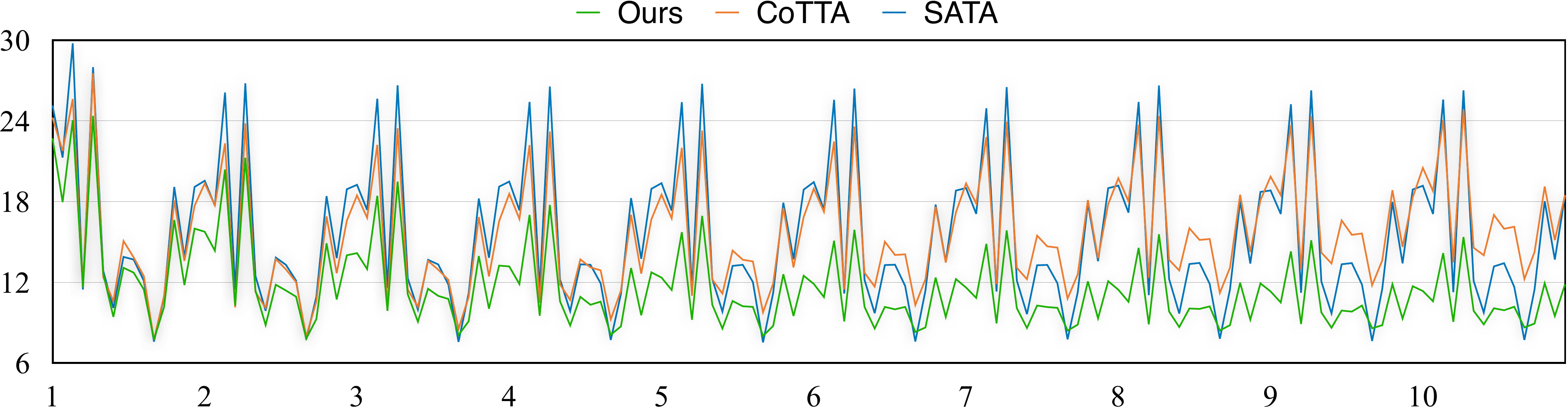}
    \caption{Under the CIFAR10-C dataset, it can be observed that the performance of each method under 10 corruption cycles varies. The error rates of CoTTA and SATA methods start to gradually increase in the later stages, whereas our method continues to decrease or maintain stability.}
    \label{fig:loops}
\end{figure}

 We present the results of 10 cycles on CIFAR10-C. As depicted in the \text{Fig.}~\ref{fig:loops}, it is evident that over time, the error rate of CoTTA and SATA methods has gradually increased, whereas our method continues to decrease. Consequently, the performance gap is widening. This phenomenon is most likely related to the fact that the first two methods lack reasonable control over the category and the offset direction of the domain.

%

\section{Hyperparameter analysis}

In this section, we delve into the critical examination of hyperparameters $\lambda_1$ and $\lambda_2$ in \text{Eq.}~\ref{eq:9} within our C-CoTTA framework in ImageNet-C, which substantially influence the model's performance. Through meticulous experimentation, we fine-tuned these hyperparameters to identify their optimal values, ensuring the harmonious interplay between Control Domain Shift (CDS) and Control Class Shift (CCS) components.

Our investigation revealed that the selection of $\lambda_1$ and $\lambda_2$ is pivotal in balancing the contributions of CDS and CCS to the overall objective function. We experimented with a spectrum of values for these hyperparameters, meticulously recording the impact on classification accuracy and domain adaptation efficacy. The empirical results, illustrated in \text{Fig.}~\ref{fig:hyper}, present a compelling case for the optimal balance that our chosen hyperparameters provide, underscoring the model's robustness against various disturbances.

\begin{minipage}{\textwidth}

\begin{minipage}[t]{0.62\textwidth}
\makeatletter\def\@captype{figure}
\includegraphics[scale=0.20]{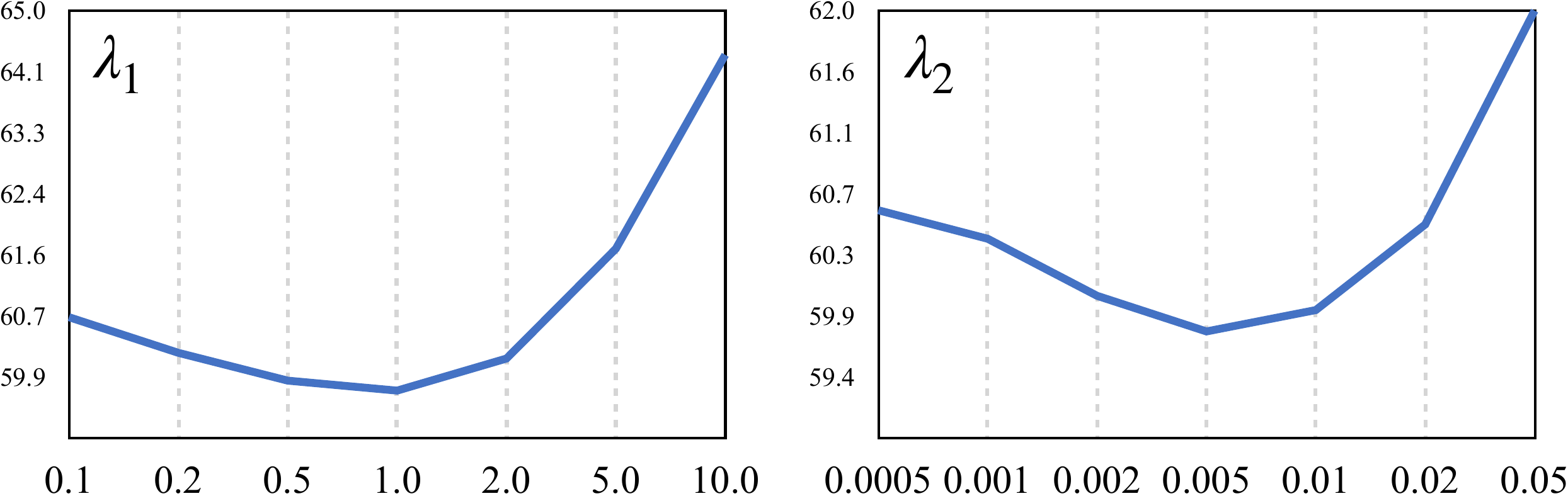}
\caption{Analysis Hyperparameter of $\lambda_1$ for CDS and $\lambda_2$ for CCS on ImageNet-C.}
\label{fig:hyper}
\end{minipage}
\hfill
\begin{minipage}[t]{0.35\textwidth}
\makeatletter\def\@captype{figure}
\includegraphics[scale=0.20]{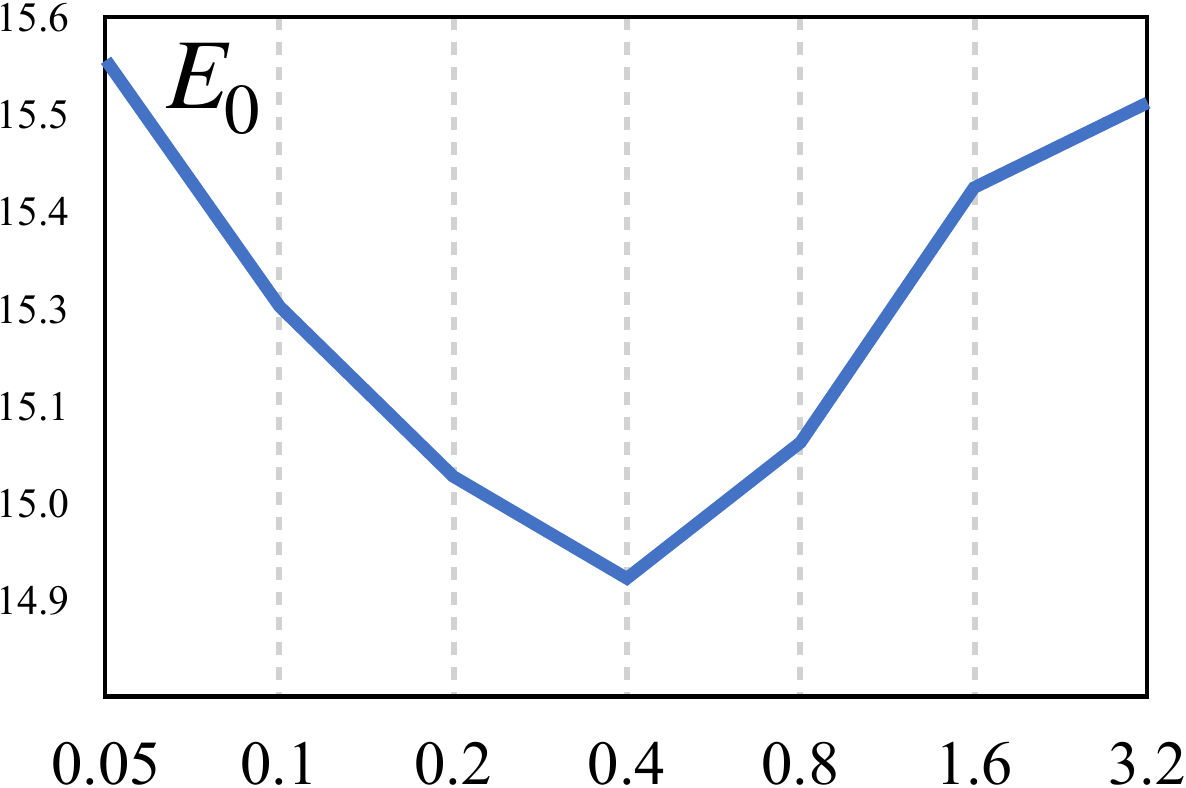}
\caption{Analysis Hyperparameter of $E_0$ on CIFAR10-C.}
\label{fig:hyper_}
\end{minipage}

\end{minipage}

\section{Reliable Sample Selection Analysis}

In our control category offset method, in order to reduce the contamination of category prototypes by misclassified samples and thus affect the control of category shift direction, we remove samples with entropy values exceeding the predefined threshold $E_0$, which is set as $0.4 \times \ln C$ based on \cite{niu2022eata}. We have verified the rationality of this operation through experiments. As shown in \text{Fig.}~\ref{fig:hyper_}, when the threshold is large, the effectiveness of the CCS method deteriorates. This may be because the conditions are too loose, leading to a large number of misclassified samples when calculating the prototype. On the other hand, when the threshold is small, the CCS method also deteriorates. This may be because the conditions are too strict, resulting in too few samples used to calculate the prototype, making the generated prototype not representative.

\section{Class Confusion Matrix}

\begin{figure}[h!]
    \setlength{\abovecaptionskip}{0pt}
    \centering
    \includegraphics[scale=0.24]{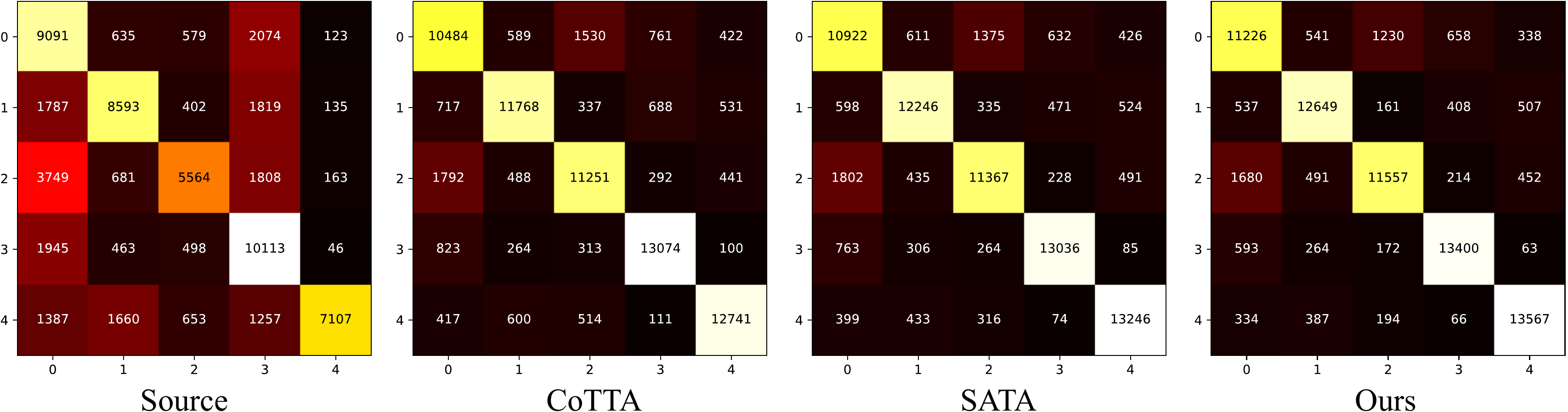}
    \caption{confusion matrix of category in the domain adaptation process. The vertical axis represents the true labels, and the horizontal axis represents the predicted labels.}
    \label{fig:confusion}
\end{figure}

We observed the confusion matrix of category in the domain adaptation process. As shown in \text{Fig.}~\ref{fig:confusion}, compared to the CoTTA and SATA methods, our method significantly reduced the degree of category confusion, demonstrating the effectiveness of controllable domain shift.

%

\end{document}